\documentclass[10pt,twocolumn,letterpaper]{article}

\usepackage{iccv}              %

\usepackage{pifont}
\usepackage{gradient-text}

\usepackage[table,xcdraw,dvipsnames]{xcolor} %
\newcommand{\gr}{\rowcolor[gray]{.92}} %
\usepackage{tabularx}
\usepackage{bm}
\usepackage{tikz}
\usepackage{multirow}
\usepackage{balance}
\usepackage[accsupp]{axessibility}  %

\definecolor{iccvblue}{rgb}{0.21,0.49,0.74}
\usepackage[pagebackref,breaklinks,colorlinks,allcolors=iccvblue]{hyperref}

\def\confName{ICCV}
\def\confYear{2025}

\newcommand{\name}{SpaRC-AD{}} %

\title{SpaRC-AD: A Baseline for Radar-Camera Fusion \\ in End-to-End Autonomous Driving}

\author{Philipp Wolters\quad %
Johannes Gilg\quad
Torben Teepe\quad
Gerhard Rigoll\quad
\\ [0.25cm]
Technical University of Munich \qquad
}

\begin{document}
\maketitle
\begin{abstract}
    \textbf{End-to-end autonomous driving} systems promise stronger performance through unified optimization of perception, motion forecasting, and planning.
    However, vision-based approaches face fundamental limitations in adverse weather conditions, partial occlusions, and precise velocity estimation -- critical challenges in safety-sensitive scenarios where accurate motion understanding and long-horizon trajectory prediction are essential for collision avoidance. 
    To address these limitations, we propose \textbf{\name{}}, a query-based end-to-end camera-radar fusion framework for planning-oriented autonomous driving.
    Through sparse 3D feature alignment and Doppler-based velocity estimation, we achieve strong 3D scene representations for refinement of agent anchors, map polylines, and motion modelling.
    Our method achieves strong improvements over the state-of-the-art vision-only baselines across multiple autonomous driving tasks, including 
    3D detection ($+4.8\%$ mAP), 
    multi-object tracking ($+8.3\%$ AMOTA), 
    online mapping ($+1.8\%$ mAP), 
    motion prediction ($-4.0\%$ mADE), and 
    trajectory planning ($-0.1$m L2 and $-9\%$ TPC).
    We achieve both spatial coherence and temporal consistency on multiple challenging benchmarks, including \textbf{real-world open-loop nuScenes}, long-horizon T-nuScenes, and \textbf{closed-loop simulator Bench2Drive}.
    We show the effectiveness of radar-based fusion in safety-critical scenarios where accurate motion understanding and long-horizon trajectory prediction are essential for collision avoidance. 
    The source code of all experiments is available at \url{https://phi-wol.github.io/sparcad/}
    
\end{abstract}

\section{Introduction}
\label{sec:intro}

\begin{figure}[t]
    \centering
    \includegraphics[width=\columnwidth]{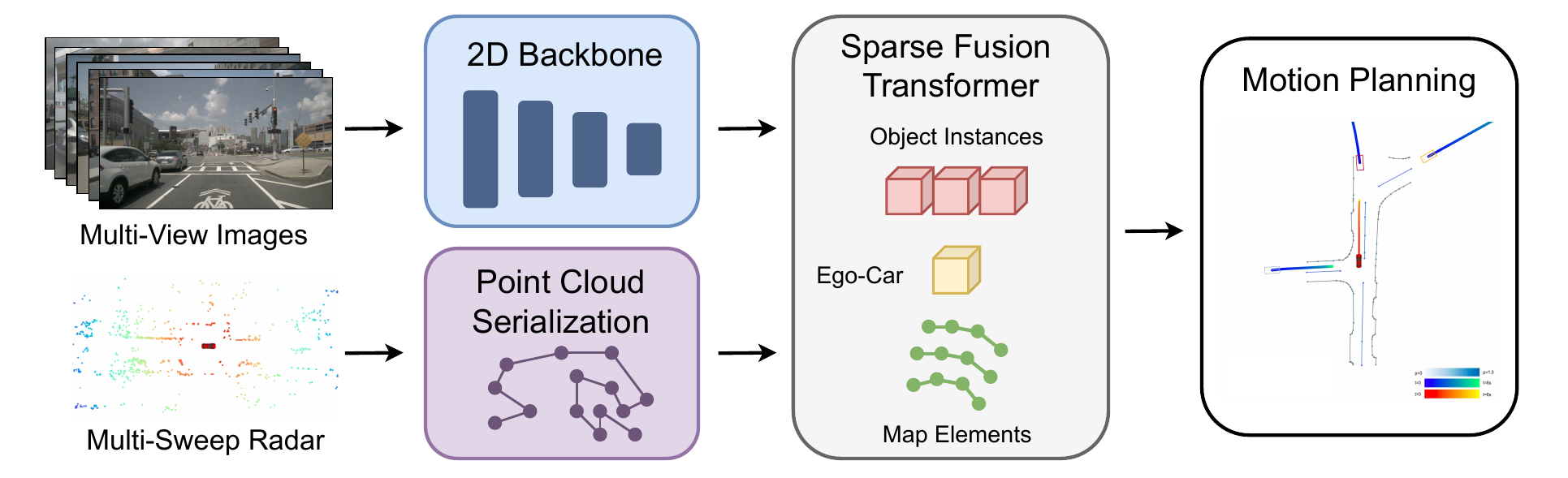}
    \caption{
        We propose \textbf{\name}, a query-based camera-radar fusion framework for autonomous driving that jointly optimizes perception, prediction and planning.
    }
    \label{fig:teaser}
\end{figure}

Autonomous driving systems have evolved from modular, multi-stage perception pipelines to unified end-to-end learning frameworks that directly map raw sensor inputs to vehicle control commands \cite{chen2024end}. 
While conventional approaches decompose the driving task into independent modules for 3D object detection \cite{huang2021bevdet, yang2023bevformer, Wang2023streampetr}, multi-object tracking \cite{zeng2022motr, zhang2021fairmot, yin2021center, teepe2024lifting}, and online mapping \cite{liu2023vectormapnet, liao2022maptr, yuan2024streammapnet}, 
recent end-to-end methods \cite{hu2023uniad, zhang2024sparsead} demonstrate the advantages of joint optimization across perception, prediction, and planning tasks. 

The new optimization objective is to generate driving controls and trajectories for the ego vehicle, directly from sensor inputs of cameras, LiDARs, and radars \cite{chitta2022transfuser}.
Leveraging expert demonstrations through imitation learning, raw sensor signals are directly processed to output vehicle motion plans and intermediate representations optimized towards the final planning goal.
Initially in Bird's Eye View (BEV) representations \cite{liu2022bevfusion}, the future trajectory of the ego vehicle is regressed from an ego-token within a transformer decoder, reducing the problem to a supervised learning setting \cite{sun2024sparsedrive}.

However, state-of-the-art research has focused on vision-centric approaches, limiting their robustness in challenging scenarios such as adverse weather conditions, partial occlusions, and long-range detection.

Critical for planning safety: robust depth estimation, strong motion-forecasting, stable trajectories. 
Song et al. have shown that especially in turning scenarios, models suffer from unstable trajectories, vulnerability to occlusions, and temporal inconsistencies \cite{song2025momad}.
The implicit depth modeling in query-based transformers lacks geometric constraints, leading to substantial localization errors in 3D perception due to unreliable depth estimation \cite{wolters2024sparc}.
Due to noise from highly dynamic environments and following detection errors, uncertainties arise in long-time horizon and long-range planning.
Moreover, causal confusion and the reliance on temporal smoothness of the ego trajectory and past motion pose a challenge \cite{li2024ego}.

Radar sensors provide critical advantages that address fundamental limitations of vision-centric approaches in end-to-end autonomous driving. 
Their robust long-range detection capabilities beyond 150m, direct velocity measurements through Doppler effects, and weather-independent operation enable more reliable spatial reasoning through time-of-flight range measurements \cite{wolters2024sparc, fent2023radargnn, zhou2023bridging}. 
Additionally, radar's ability to measure relative velocities enhances multi-agent intent prediction, leading to more stable and consistent trajectory planning. 
These complementary strengths make radar fusion particularly valuable for safety-critical autonomous driving applications.

While multi-modal fusion with cameras and LiDAR has shown benefits \cite{chitta2022transfuser}, and radar fusion has proven effective for modular perception \cite{lin2024rcbevdet, wolters2024hydra, chu2025racformer}, the integration of radar into end-to-end autonomous driving remains unexplored.
We investigate the impact and potential of including radar into the end-to-end optimization and how to leverage the additional motion cues reflected from the environment.
Due to sparsity of the radar representation and precise spatial-temporal calibration, we propose a query-based approach that iteratively refines the motion and positional charactersitcs of map and traffic agent representations.

In this work, we address the critical gap in radar-based end-to-end autonomous driving by proposing \textbf{\name}{}, extending the sparse representation paradigm of radar points and scene instances in a coherent end-to-end framework, and creating synergies between radar data characteristics and planning requirements.
Our approach iteratively refines motion and positional characteristics of both map and agent representations by leveraging spatial proximity of reflected radar points as strong inductive biases.

Our main contributions are:
\begin{itemize}[leftmargin=+.5cm]
    \item First radar-based end-to-end autonomous driving baseline on key benchmarks.
    \item Extension of sparse fusion design for simultaneous detection, tracking, and planning queries.
    \item Holistic radar-based fusion improves 3D detection (+4.8\% mAP), multi-object tracking (+8.3\% AMOTA), online mapping (+1.8\% mAP), and motion forecasting (-4.0\% mADE), optimizing trajectory prediction consistency (-9.0\% TPC) and simulation success rates~(+10.0\%).
    \item Extensive evaluation on multiple benchmarks of open-loop nuScenes \cite{caesar2020nuscenes} and closed-loop simulation of Bench2Drive \cite{jia2024bench2drive}.
    \item We provide additional qualitative analysis demonstrating superior performance through enhanced perception range, more accurate motion modeling, and increased robustness under challenging environmental conditions.
\end{itemize}

\section{Related Work }
\label{sec:related_work}

\subsection{Planning Oriented Autonomous Driving}

A new paradigm has emerged in autonomous driving research, moving from multi-stage frameworks \cite{li2023delving, jia2023towards, li2024think2drive} to end-to-end autonomous driving \cite{chen2024end}.
This evolution addresses the fundamental limitations of modular approaches: information loss and error accumulation across subsequent stages, which constrain optimal system performance.
The goal is to strengthen generalization to complex driving scenarios in a data-driven manner.

Typically the state-of-the-art methods follow an encoder-decoder principle, first encoding the sensor data into a latent representation, then decoding the intermediate representation into a driving policy \cite{chitta2022transfuser, hu2022st}.
The pioneering works of UniAD \cite{hu2023uniad} and VAD \cite{jiang2023vad, chen2024vadv2} have shown that all tasks are communicated within unified query interfaces, enabling goal-oriented optimization through vectorized scene representations.
VADv2 \cite{chen2024vadv2} extends the planner to probabilistic planning, while Hydra-MDP \cite{li2024hydra} integrates additional supervision from rule-based planning modules.
SparseDrive \cite{sun2024sparsedrive} explores sparse scene representations for efficient scene modeling, discarding Birds-Eye-View (BEV) representations.

\subsection{Camera-Radar 3D Perception}

In 3D object detection, radar-camera-based approaches have emerged as a low-cost and robust alternative to lidar-based perception.
Initial works fused in the perspective view \cite{nabati2021centerfusion, kim2022craft, nobis19crfnet, long2023radiant}, associating the sparsely projected radarpoints to the dense encoded image features.

Grid-rendering approaches have adapted the BEVFusion \cite{liu2022bevfusion} paradigm to the characteristics of radar sensors \cite{wolters2024hydra, kim2023crn, lin2024rcbevdet, kim2024crt, kim2024rcm, ronecker2024dynamic, ronecker2024deep}, by
encoding them with PointPillars \cite{lang2019pointpillars} or Voxels \cite{zhou2018voxelnet}, that are densely parametrized, but sparse in information density. The feature maps are then combined in BEV space.
CRN \cite{kim2023crn}, HyDRa \cite{wolters2024hydra}, and RCBEVDet \cite{lin2024rcbevdet} tackle the spatial misalignment between radar and camera sensors, surpassing vision-based approaches with stronger velocity prediction, depth estimation, and robustness in adverse weather conditions.

While RaCFormer \cite{chu2025racformer} still utilizes BEV-encoded radar features but decodes the features via sampling in a transformer, SpaRC \cite{wolters2024sparc} proposes a new state-of-the-art in 3D object detection via fully sparse encoding and fusion of radar points. 
Through point cloud serialization in the backbone, it enables a direct point-to-object interaction, dynamically weighted with strong priors for the subsequent perspective aggregation and hierarchical query optimization. 
We will leverage these principles to design a query-based fusion of radar points and scene instances of the full surrounding environment, by reducing the spatial and temporal uncertainty, to facilitate the planning task.

\begin{figure*}[t]
    \centering
    \includegraphics[width=0.82\textwidth]{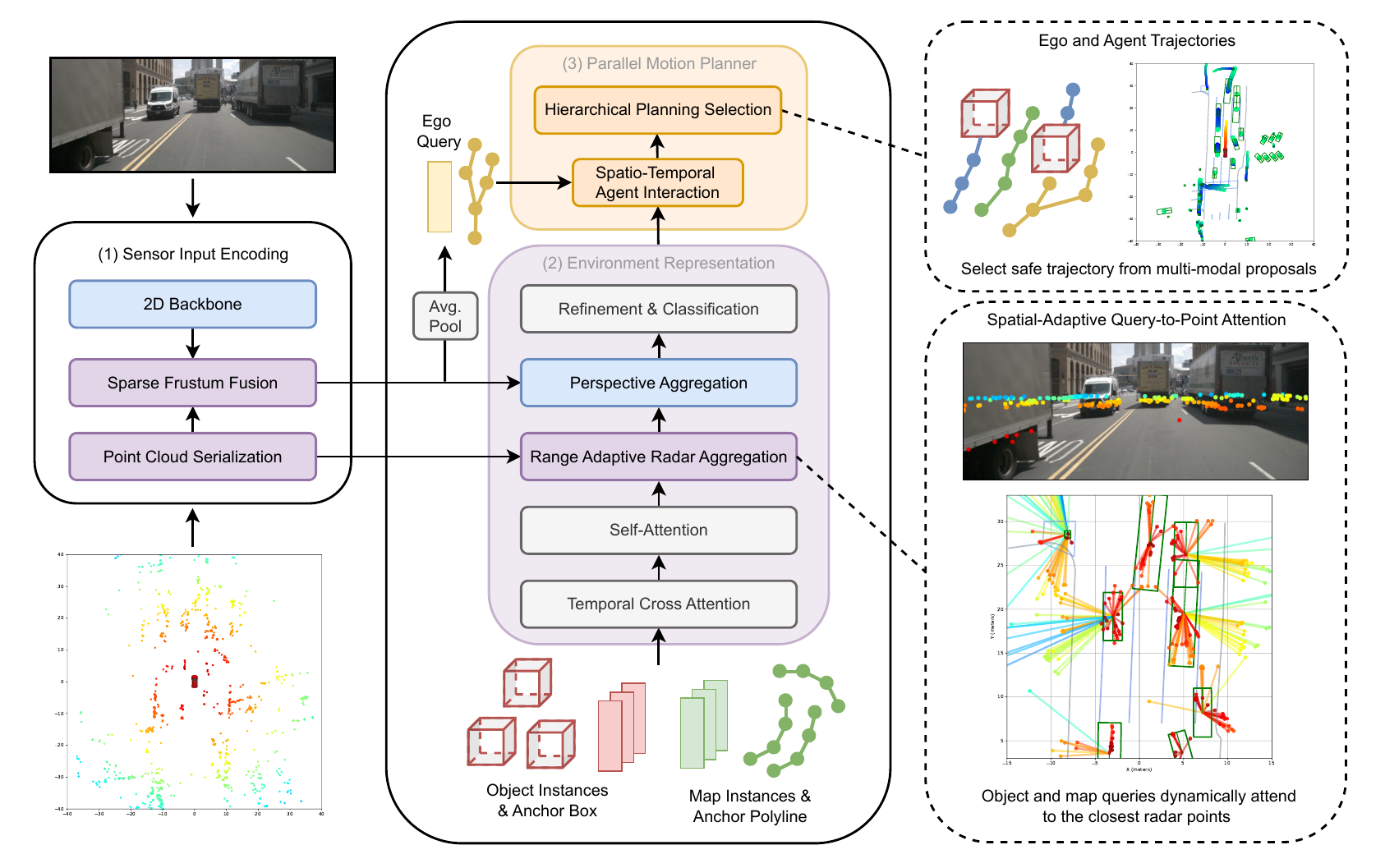}
    \caption{
        \textbf{Overview of \name}.
        Our framework consists of three main components: 
        (1)~multi-modal sparse feature encoding that processes camera and radar inputs, 
        (2)~unified sparse fusion that leverages query-based point interactions between modalities, and 
        (3)~parallel motion planning that jointly optimizes the strengthened spatial scene representations for trajectory generation.
        }
    \label{fig:overview}
\end{figure*}
\section{Architecture}
\label{sec:main}

\subsection{Framework Overview.}

\name{} extends the sparse-centric transformer of SparseDrive \cite{sun2024sparsedrive} by integrating the adaptive radar fusion strategies from SpaRC \cite{wolters2024sparc} into a unified end-to-end autonomous driving framework. 
Our approach addresses the fundamental challenge of fusing radar representations with dense visual features in a planning-oriented optimization pipeline.

The overall architecture consists of three main components: 
(1) multi-modal sparse feature encoding that processes camera and radar inputs into compatible representations, 
(2) unified sparse fusion that leverages query-based interactions between modalities, and 
(3) parallel motion planning that jointly optimizes the strengthened spatial scene representations for perception, prediction, and trajectory generation.
This design enables direct end-to-end fusion and optimization without leveraging inefficient grid-based representations.

Our framework processes 360-degree surround-view images through a 2D convolutional neural network backbone with a feature pyramid neck,
generating multi-view multi-scale feature maps.
Simultaneously, multi-sweep radar point clouds (spatial coordinates, RCS intensity, and Doppler velocity) are encoded into sparse feature representations through point-wise encoding and serialization using Point Transformer \cite{wu2024point}, producing a set of 3D embedded radar features.

\begin{table*}[htpb]
    \centering
    \resizebox{\textwidth}{!}{
    \begin{tabular}{l|cc|cccc|cccc|cccc}
    \toprule
    \multirow{2}{*}{\textbf{Method}} & \multirow{2}{*}{\textbf{Input}} & \multirow{2}{*}{\textbf{Backbone}} &
    \multicolumn{4}{c|}{\textbf{L2 (m) $\downarrow$}} & 
    \multicolumn{4}{c|}{\textbf{Col. Rate (\%) $\downarrow$}} & 
    \multicolumn{4}{c}{\textbf{TPC (m) $\downarrow$}} \\
    && & 1s & 2s & 3s & \cellcolor{gray!15}Avg. & 1s & 2s & 3s & \cellcolor{gray!15}Avg. & 1s & 2s & 3s & \cellcolor{gray!15}Avg. \\
    \midrule
    UniAD~\cite{hu2023uniad} & C & R101  
    & 0.45&0.70&1.04 & \cellcolor{gray!15}0.73 
    & 0.62&0.58&0.63 & \cellcolor{gray!15}0.61 & 0.41 &0.68 &0.97 
    & \cellcolor{gray!15}0.68 \\ %
    
    VAD~\cite{jiang2023vad}  & C & R50  
    & 0.41 & 0.70&1.05 & \cellcolor{gray!15}0.72 
    & 0.07&0.17&0.41 & \cellcolor{gray!15}0.22 & 0.36 & 0.66 & 0.91 
    & \cellcolor{gray!15}0.64 \\

    GenAD \cite{zheng2024genad} & C & R50 
    & 0.28 & 0.49 & 0.78 & \cellcolor{gray!15}0.52 
    & 0.08 & 0.14 & 0.34 & \cellcolor{gray!15}0.19 
    & - & - & - & \cellcolor{gray!15}- \\ %

    MomAD \cite{song2025momad} & C & R50 
    & 0.31 & 0.57 & 0.91 & \cellcolor{gray!15}0.60 
    & 0.01 & 0.05 & 0.22 & \cellcolor{gray!15}0.09  
    & 0.30 & 0.53 & 0.78 & \cellcolor{gray!15} 0.54 \\ %
    
    BridgeAD \cite{zhang2025bridgead} & C & R50 
    & 0.29 & 0.57 & 0.92 & \cellcolor{gray!15}0.59 
    & 0.01 & 0.05 & 0.22 & \cellcolor{gray!15}0.09 
    & - & - & - & \cellcolor{gray!15}- \\ %
    
    DiffusionDrive~\cite{liao2025diffusiondrive} & C 
    & R50 & 0.27 & 0.54 & 0.90 & \cellcolor{gray!15}0.57 
    & 0.03 & 0.05 & 0.16 & \cellcolor{gray!15} \textbf{0.08} 
    & - & - & - & \cellcolor{gray!15} - \\ %
    
    \midrule
    SparseDrive~\cite{sun2024sparsedrive}& C & R50 
    & 0.29 & 0.58 & 0.96 & \cellcolor{gray!15}0.61 
    & 0.01 & 0.05 & 0.18 & \cellcolor{gray!15} \textbf{0.08} 
    & 0.30 & 0.57 & 0.85 & \cellcolor{gray!15}0.57 \\ %
    
    \gr \textbf{\name} (Ours) & C+R & R50 
    & \textbf{0.24} & \textbf{0.47} &  \textbf{0.79} & \textbf{0.50} & 0.01 &  0.06 & 0.20 & 0.09 &\textbf{0.27} & \textbf{0.47} & \textbf{0.70} & \textbf{0.48} \\
    \bottomrule
\end{tabular}%

    }
    \caption{Comparison on \textbf{nuScenes} dataset with \textbf{open-loop} metrics.
    Metric calculation follows VAD \cite{jiang2023vad} and MomAD \cite{song2025momad}.
    C and R denote Camera and Radar. 
    Similar to SparseDrive \cite{sun2024sparsedrive} and MomAD \cite{song2025momad}, we deactivate the ego status information for a fair comparison (preventing ego status leakage as analyzed in\cite{li2024ego}).
    }
    \label{table:nusc_planning}
\end{table*}

\subsection{Query Design}

Detection queries represent surrounding traffic agents as anchor boxes with eleven parameters: ${x, y, z, w, h, l, \sin \theta, \cos \theta, v_x, v_y, v_z}$, where spatial coordinates, dimensions, orientation, and velocity are jointly predicted and optimized. 
These anchors $\mathbf{B}_d \in \mathbb{R}^{N_d \times 11}$ are paired with instance features $\mathbf{F}_d \in \mathbb{R}^{N_d \times C}$ obtained through K-means clustering on the training set.

Map element queries model static road infrastructure as polylines with $N_p$ waypoints: ${x_0, y_0, x_1, y_1, \ldots, x_{N_p-1}, y_{N_p-1}}$. 
Map instances are represented by features $\mathbf{F}_m \in \mathbb{R}^{N_m \times C}$ and anchor polylines $\mathbf{L}_m \in \mathbb{R}^{N_m \times N_p \times 2}$, with each element containing up to 20 waypoints.

\subsection{Sparse Fusion}

Following SpaRC's design \cite{wolters2024sparc}, we implement range-adaptive aggregation that dynamically weights radar features based on their spatial proximity to query locations. 
We aggregate nearby radar features for each query instance using distance-weighted attention that dynamically adjusts feature importance based on spatial proximity:
\begin{equation}\label{rara}
    \text{Attn}(\mathbf{q}, \mathbf{k}, \mathbf{v}) = \text{softmax}\left(\frac{\mathbf{q}\mathbf{k}^T}{\sqrt{d}} - \alpha \frac{\|\mathbf{p}_q - \mathbf{p}_k\|_2}{r_\text{max}}\right)\mathbf{v}
\end{equation}
where $\mathbf{q} \in \mathbb{R}^{N_q \times d}$ queries attend to radar key-value pairs $\mathbf{k}, \mathbf{v} \in \mathbb{R}^{N_k \times d}$ via scaled dot-product attention with a distance-based penalty term. 
The 3D positions $\mathbf{p}_q \in \mathbb{R}^{N_q \times 3}$ and $\mathbf{p}_k \in \mathbb{R}^{N_k \times 3}$ are normalized by $r_\text{max}$.

For map elements, we compute the minimum distance between a radar point and polyline segments:

\begin{equation}\label{eq:line_dist}
    d_{\text{min}} = \min_{i=1}^{N_p-1} \left\| \mathbf{p}_r - \left(\mathbf{p}_i + t \cdot (\mathbf{p}_{i+1} - \mathbf{p}_i)\right) \right\|_2
\end{equation}
where $\mathbf{p}_r$ is the radar point position, $\mathbf{p}_i$ and $\mathbf{p}_{i+1}$ are consecutive polyline points, and $t$ is the projection parameter clamped between 0 and 1. The projection parameter $t$ is computed as:
\begin{equation}
    t = \text{clamp}\left(\frac{(\mathbf{p}_r - \mathbf{p}_i) \cdot (\mathbf{p}_{i+1} - \mathbf{p}_i)}{\|\mathbf{p}_{i+1} - \mathbf{p}_i\|_2^2}, 0, 1\right)
\end{equation}

This distance metric enables effective attention between radar points and map elements by considering the closest line segment of each polyline.
After radar-based set-to-set aggregation, the decoder module encompasses iterative blocks of deformable perspective aggregation \cite{lin2022sparse4d}, self-attention, and feedforward networks. 
While the deformable aggregations uses learnable keypoints around the anchor boxes, the radar module dynamically aggregates the closest radar features in the vicinity of the anchor boxes and polyline.

\subsection{Multi-Modal Perspective Feature Maps}

To align multi-modal features across perspectives and 3D representations, we additionally employ sparse frustum fusion \cite{wolters2024sparc} that projects radar points into camera frustums and performs cross-attention between radar features and image regions.
Thus, the ego-vehicle instance benefits directly from the radar-enriched representation, when Average Pooling the feature representation into a single query initialization.

This provides the ego instance with rich semantic and geometric information essential for planning-oriented optimization, incorporating both visual context and radar-derived motion cues.

\subsection{Probabilistic Trajectory Modeling}

On top of the fusion representation, we leverage agent-level interactions via cross-attention, fusing history information of the agents and map elements.
Each query, including the ego-instance predicts multi-modal trajectories following the three driving commands: turn left, turn right, and go straight.
Each trajectory gets re-scored based on the proximity to other agents' trajectories \cite{sun2024sparsedrive}.

\subsection{Loss Design}
The final loss function is the average displacement error~(ADE) between output and ground truth trajectories of the planned ego vehicle and the foretasted surrounding traffic agents.
Focal loss handles the classification of the trajectory modes (lowest ADE corresponds to the positive sample, others as negative samples) and L1 loss supervises the actual trajectory. 
The queries are regularized by detection and mapping loss through Hungarian matching and box/point regression losses.
A depth head in the perspective view guides with an additional L1 loss. 

The unified architecture enables joint optimization of radar fusion and planning objectives, resulting in improved spatial coherence, temporal consistency, and collision avoidance compared to vision-only baselines.

\section{Experiments}
\label{sec:experiments}
\begin{table*}[htpb]
    \Large
    \centering
      
      \resizebox{\textwidth}{!}{
      \begin{tabular}{l|ccccccc|ccc|cccc|cccc}
    \toprule
    \multirow{2}{*}{\textbf{Method}} & \multicolumn{7}{c|}{\textbf{3D Object Detection}} & \multicolumn{3}{c|}{\textbf{Multi-Object Tracking}} & \multicolumn{4}{c|}{\textbf{Online Mapping}} & \multicolumn{4}{c}{\textbf{Motion Prediction}} \\
    & mAP$\uparrow$ & NDS$\uparrow$ & mATE$\downarrow$ & mASE$\downarrow$ & mAOE$\downarrow$ & mAVE$\downarrow$ & mAAE$\downarrow$ 
    & AMOTA$\uparrow$ & AMOTP$\downarrow$ & Recall$\uparrow$ 
    & mAP$\uparrow$ & $\operatorname{AP}_{\operatorname{ped}}\uparrow$ & $\operatorname{AP}_{d}\uparrow$ & $\operatorname{AP}_{b}\uparrow$ 
    & mADE$\downarrow$ & mFDE$\downarrow$ & MR$\downarrow$ & EPA$\uparrow$ \\
    \midrule
    UniAD \cite{hu2023uniad} & \cellcolor{gray!15}38.0 & \cellcolor{gray!15}49.8 & 0.684 & 0.277 & 0.383 & 0.381 & 0.192 
    & \cellcolor{gray!15}0.359 & 1.320 & 0.467 
    & \cellcolor{gray!15}- & - & - & - 
    & \cellcolor{gray!15}0.71 & 1.02 & 0.151 & 0.456 \\
    
    VAD \cite{jiang2023vad} & \cellcolor{gray!15}31.2 & \cellcolor{gray!15}43.5 & 0.610 & 0.288 & 0.541 & 0.534 & 0.228 
    & \cellcolor{gray!15}- & - & - 
    & \cellcolor{gray!15}47.6 & 40.6 & 51.5 & 50.6 
    & \cellcolor{gray!15}- & - & - & - \\
    
    MomAD \cite{song2025momad} &  \cellcolor{gray!15}42.3 &  \cellcolor{gray!15}53.1 & 0.561 & 0.269 & 0.549 & 0.258 & 0.188 
    & \cellcolor{gray!15}0.391 & 1.243 & 0.509 
    & \cellcolor{gray!15}55.9 & 50.7 & \textbf{58.1} &58.9
    & \cellcolor{gray!15}0.61 & 0.98 & 0.137 & 0.499 \\
    \midrule
    SparseDrive \cite{sun2024sparsedrive} & \cellcolor{gray!15}41.8 & \cellcolor{gray!15}52.5 & 0.566 & 0.275 & 0.552 & 0.261 & 0.190 
    & \cellcolor{gray!15}0.386 & 1.254 & 0.499 
    & \cellcolor{gray!15}55.1 & 49.9 & 57.0 & 58.4 
    & \cellcolor{gray!15}0.62 & 0.99 & 0.136 & 0.482 \\
    \gr \textbf{\name} (Ours) & \cellcolor{gray!15} \textbf{46.6} & \cellcolor{gray!15} \textbf{57.0} & \textbf{0.512} & \textbf{0.271} & \textbf{0.494} & \textbf{0.173} & \textbf{0.177} 
    & \cellcolor{gray!15} \textbf{0.469} & \textbf{1.129} & \textbf{0.553} 
    & \cellcolor{gray!15} \textbf{56.9} & \textbf{53.7} & 55.4 & \textbf{61.7}
    & \cellcolor{gray!15} \textbf{0.58} & \textbf{0.93} & \textbf{0.121} & \textbf{0.53} \\

    \bottomrule
    
    \end{tabular} }
    \caption{Perception and motion results on the \textbf{nuScenes} validation dataset. $\operatorname{AP}_{d}$ denotes $\operatorname{AP}_{\operatorname{divider}}$. $\operatorname{AP}_{b}$ denotes $\operatorname{AP}_{\operatorname{boundary}}$. $\operatorname{mADE}$ denotes $\operatorname{minADE}$. $\operatorname{mFDE}$ denotes $\operatorname{minFDE}$.}
    \label{table:nusc_perception}
    \end{table*}

\subsection{Experimental Setup}

For a comprehensive evaluation, we evaluate our approach on real-world open-loop benchmarks as well as a closed-loop simulation environment.

\noindent\textbf{nuScenes Open-Loop} \cite{caesar2020nuscenes}
We evaluate on the standard nuScenes dataset containing 1000 driving scenes of 20 seconds each at 2Hz, captured by six surround-view cameras, one LiDAR and 5 radars, collecting point clouds including RCS and Doppler velocity measurements. 

\noindent\textbf{Long-Horizon Turning-nuScenes} \cite{song2025momad}
To better assess the temporal consistency of predicted trajectories, Song et al. introduced a new validation benchmark based on the most challenging turning scenarios within the nuScenes validation set.

\noindent\textbf{Bench2Drive} \cite{jia2024bench2drive}
The NeurIPS 2024 benchmark is a reactive simulation environment for autonomous driving following a closed-loop evaluation protocol under CARLA Leaderboard 2.0 \cite{Jaeger2023ICCV}. 
We use the official base configuration of 1000 simulated driving scenes, captured by six surround-view cameras and 5 radar sensors collecting sparse point clouds with velocity measurements.
The sensor setup closely resembles the vehicle configuration of nuScenes.
We evaluate on the dev10 protocol \cite{jia2025drivetransformer}, an officially curated subset of varying weather conditions, locations, and traffic densities selected to cover a wide range of difficult driving scenarios with low variance.

\noindent\textbf{Evaluation Metrics}
We follow the established evaluation protocols for comprehensive assessment across all autonomous driving tasks:
3D Object Detection: Average precision (mAP) and nuScenes Detection Score (NDS), which comprises the weighted sum of mAP and five True Positive metrics: Translation (mATE), Scale (mASE), Orientation (mAOE), Velocity (mAVE), and Attribute Error (mAAE).
Multi-Object Tracking: Average Multi-Object Tracking Accuracy (AMOTA) and Average Multi-Object Tracking Precision (AMOTP).
Online Mapping: Map segmentation accuracy using mean Average Precision (mAP) for different map elements, including pedestrian crossings ($AP_{ped}$), lane dividers ($AP_{d}$), and lane boundaries ($AP_{b}$).
Motion Prediction:Minimum Average Displacement Error (minADE), minimum Final Displacement Error (minFDE), Miss Rate (MR), and End-to-end Prediction Accuracy (EPA) \cite{hu2023uniad}.
Planning: L2 Displacement Error (L2), Collision Rate and Trajectory Prediction Consistency (TPC) \cite{song2025momad}.
For all planning metrics, we are following \cite{sun2024sparsedrive, song2025momad} which follow the official settings introduced by VAD \cite{jiang2023vad}.
During reactive closed-loop evaluation, we additionally evaluate the Bench2Drive driving score and the success rate of the planned trajectories.

\subsection{Implementation Details}

We follow the multi-stage training pipeline of \cite{sun2024sparsedrive}.
In the first stage, we train the multi-modal sparse feature encoder and the detection head. 
Each modality backbone is trained from scratch (ResNet initialized from an ImageNet checkpoint).

\name{} uses a single configuration of 900 anchors for detection, 100 polylines for mapping, and 6 decoder layers.
We employ the AdamW optimizer and Cosine Annealing learning rate scheduler for 100 epochs (similar to \cite{sun2024sparsedrive} and \cite{song2025momad}) in stage one and 10 epochs in stage two.
Further hyper-parameters will be provided in the accompanying code repository.

The perception range is set to 50m, with an instance memory queue of three key frames, training in a streaming manner \cite{Wang2023streampetr}.
The motion forecasting horizon is set to 12s and the planning prediction to 6s.
The vison backbone encompasses a ResNet-50 with an input-size of $256\times704$ on nuScenes and $384\times704$ on Bench2Drive (same as all compared model configurations).
Our models are trained with a batch size of 48.

We deactivate ego status information following SparseDrive conventions \cite{sun2024sparsedrive, song2025momad} to prevent ego status leakage as analyzed in \cite{li2024ego}, ensuring fair comparison across all methods.

\subsection{Main Results}
\subsubsection{Perception and Motion Forecasting Results}

As shown in \cref{table:nusc_perception}, \name{} achieves significant improvements across all perception tasks compared to the SparseDrive baseline. 
Our radar fusion framework demonstrates a 4.8\% mAP improvement and 4.5 NDS enhancement on the nuScenes validation set. 
The improvements are particularly pronounced in velocity estimation (mAVE: 0.173 vs 0.261), highlighting radar's effective contribution through Doppler measurements. 

Moreover, \name{} achieves state-of-the-art tracking performance with 8.3\% AMOTA improvement over vision-centric SparseDrive.
The enhanced velocity estimation from radar Doppler directly benefits object-level motion modeling, leading to more stable tracking trajectories. 
Combined with improved precision, our approach demonstrates superior capability in maintaining object identity across frames, which is critical for planning-oriented autonomous driving systems.

The radar fusion also provides 1.8\% mAP improvement in online mapping, with particularly strong gains in lane boundary detection.
Finally, \name{} achieves a 4.0\% reduction in mADE, demonstrating improved motion forecasting accuracy. Integrating radar-derived velocity information enhances multi-agent intent prediction, leading to more accurate trajectory forecasts.

\begin{figure*}[t]
    \centering
    \includegraphics[width=0.97\textwidth]{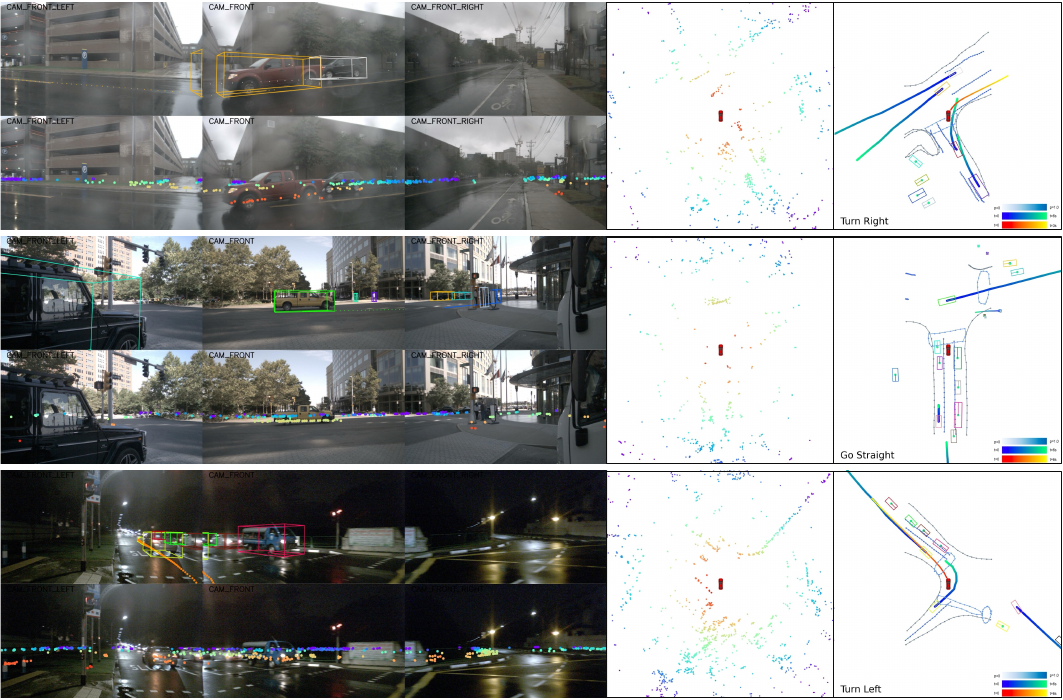}
    \caption{
        \textbf{Qualitative Examples} of produced trajectories and visualized radar points of a challenging turning scenario with a long horizon of six seconds (top: rain, middle: partially hidden objects, bottom: night). 
        On the left, we show the front-facing camera views, predicted bounding boxes, and projected radar points.
        In the middle, we visualize the perceived radar points in a top-down Bird's-Eye-View at 50m range, color-coded by the distance to the ego vehicle.
        On the right: the corresponding predicted map elements, bounding boxes with motion forecasts, and planned trajectories.
        }
    \label{fig:qualitative}
\end{figure*}

\subsubsection{Open-Loop Planning Results}

\cref{table:nusc_planning} evaluates the performance of \name{} in open-loop planning settings, with the lowest average L2 error (0.50m) compared to SparseDrive (0.61m), UniAD (0.73m), and MomAD (0.60m).
Most significantly, we achieve a 9\% improvement in Trajectory Prediction Consistency (TPC) compared to SparseDrive, indicating more consistent trajectory prediction.

In summary, \name{} achieves state-of-the-art performance on the nuScenes open-loop benchmark, demonstrating the effectiveness of radar fusion in improving perception, tracking, and motion forecasting capabilities.
The raw strength in feature representation also outperforms more sophisticated planner like MomAD \cite{song2025momad} or DiffusionDrive~\cite{liao2025diffusiondrive}. 

\subsubsection{Turning Scenarios}
When evaluating the most complex and challenging scenarios (\cf. \cref{table:nusc_t}), the difference to vision-based models increases.
We are able to significantly improve the L2~(-0.26m) and TPC metrics (-0.15), while maintaining the overall low collision rate (-31\%) of 0.09, in contrast to SparseDrive.
This safety-critical scenario analysis shows the effectiveness of our radar-based approach and emphasizes the importance of multi-modal sensor integration for all autonomous driving designs.

\subsubsection{Long Trajectory Prediction}
In \cref{table:nusc_long6}, we increase the prediction horizon to 6s and evaluate the performance of \name{} in long-term trajectory prediction.
In both settings, full-set and T-nuScenes, we are able to significantly improve the trajectory consistencies in L2 and TPC, with strongly reduced collision rates.
We can show that doubling the prediction horizon and overcoming partial occlusions in highly dynamic scenes 
shows a major potential for trajectory consistency and collision reduction.
In a six second prediction horizon, we can see that the radar-based approach is able to predict more stable trajectories.
\name{} can capitalize on longer perception ranges, detecting partially occluded objects and better motion modeling.

\subsubsection{Closed-Loop Planning Results}

In \cref{table:b2d_dev10}, we generalize the findings of \name{} to the closed-loop planning setting of Bench2Drive.
Evaluating in open-loop, we again outperform the baseline SparseDrive by a trajectory displacement of 0.82 vs 0.87m.
Moreover, in interactive scenarios like cut-ins, overtaking maneuvers, or emergency braking, \name{} achieves a 20\% higher success rate compared to SparseDrive.

\subsection{Qualitative Analysis}

Furhermore, we visualize the perception and planning performance of our model in challenging scenarios.
\cref{fig:qualitative} shows the perception and planning performance of our model visually in a challenging turning scenario with a long horizon of six seconds. 
We project the radar points onto the Bird's-Eye-View and front-facing camera views and visualize the predicted map elements, bounding boxes with motion forecasts and planned trajectories.

\begin{figure*}[t]
    \centering
    \includegraphics[width=0.9\textwidth]{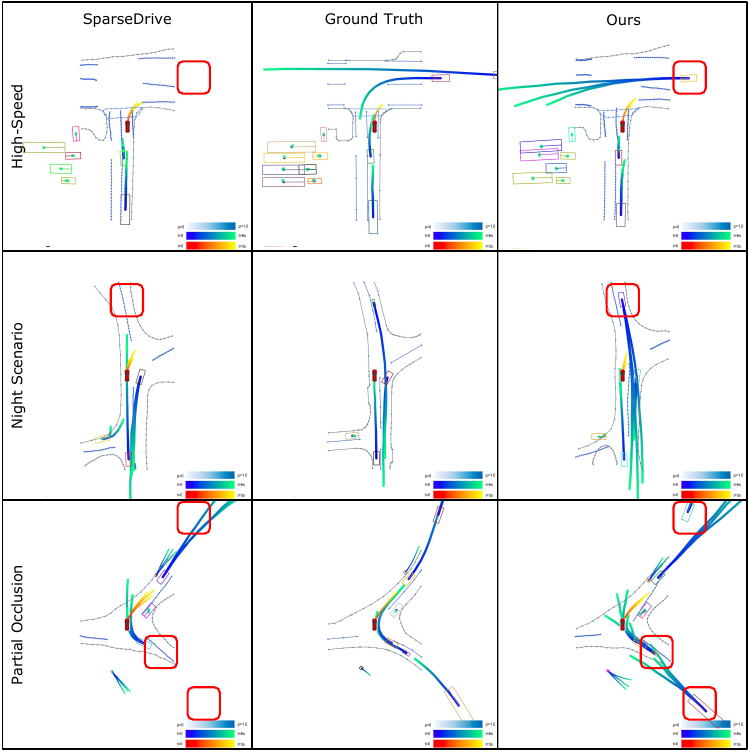}
    \caption{
        \textbf{Qualitative Comparison} of \name{} with SparseDrive on challenging turning scenarios in a crowded city environment (top-3 multi-mode trajectories).
        The first row shows a T-crossing scenario, where \name{} successfully detects an oncoming vehicle at high speed. 
        The second row shows a night scenario, where the vision baseline does not detect the oncoming scooter, whereas our approach correctly forecasts the trajectory of the camouflaged vehicle.
        The last scene emphasizes a dynamic turning scenario, where the radar-based approach is able to detect partially occluded vehicles at long range.
        }
    \vspace{-0.2cm}
    \label{fig:comparison}
\end{figure*}

In \cref{fig:comparison}, we compare our fusion design with the baseline SparseDrive and indicate, the synergies radar-fusion provides.
The qualitative analysis validates that our radar fusion strategy addresses fundamental limitations of vision-centric approaches, particularly in scenarios where precise motion understanding and long-horizon prediction are essential for collision avoidance and safe autonomous driving operation.

\begin{table}[htpb]
    \small
    \centering
      \renewcommand\arraystretch{1.0}
      \setlength{\tabcolsep}{1.5mm} %
      \resizebox{\linewidth}{!}{
      \begin{tabular}{l|cccc|cccc|cccc}
    \toprule
    \multirow{2}{*}{\textbf{Method}} & \multicolumn{4}{c}{\textbf{L2 (m) $\downarrow$}} & \multicolumn{4}{c}{\textbf{Col. Rate (\%) $\downarrow$}} & \multicolumn{4}{c}{\textbf{TPC (m) $\downarrow$}} \\
    & 1s & 2s & 3s & \cellcolor{gray!15}Avg. & 1s & 2s & 3s & \cellcolor{gray!15}Avg. & 1s & 2s & 3s & \cellcolor{gray!15}Avg. \\
    \midrule
    $\operatorname{SparseDrive}$~\cite{sun2024sparsedrive} & 0.35 & 0.77 & 1.46 & \cellcolor{gray!15}0.86 & 0.04 & 0.17 & 0.98 & \cellcolor{gray!15}0.40 & 0.34 & 0.70 & 1.33 & \cellcolor{gray!15}0.79 \\

    \gr \textbf{\name} (Ours) & 
    \textbf{0.26} & 
    \textbf{0.54} & 
    \textbf{0.93} & 
    \cellcolor{gray!15} \textbf{0.58} & 
    \textbf{0.00} & 
    \textbf{0.04} & 
    \textbf{0.23} & \cellcolor{gray!15} \textbf{0.09} & 
    \textbf{0.35} & \textbf{0.63} & \textbf{0.95} & \textbf{0.64} \\
    
    \bottomrule
    \end{tabular}}
    \caption{Planning results on the \textbf{Turning-nuScenes} validation dataset.
    We follow the VAD~\cite{jiang2023vad} evaluation metric.}
    \label{table:nusc_t}
\end{table}

\begin{table}[htp]
    \centering
    \renewcommand\arraystretch{1.0}
    \resizebox{\linewidth}{!}{%
    \begin{tabular}{ll|ccc|ccc|ccc}
    \toprule
    \multirow{2}{*}{\textbf{Split}} & \multirow{2}{*}{\textbf{Method}} & \multicolumn{3}{c|}{\textbf{L2 (m) $\downarrow$}} & \multicolumn{3}{c|}{\textbf{Col. Rate (\%) $\downarrow$}} & \multicolumn{3}{c}{\textbf{TPC (m) $\downarrow$}} \\
    & & 4s & 5s & 6s & 4s & 5s & 6s & 4s & 5s & 6s \\
    \midrule
    \multirow{2}{*}{$\operatorname{nuScenes}$} 
    & $\operatorname{SparseDrive}$~\cite{sun2024sparsedrive} & 1.75 & 2.32 & 2.95 & 0.87 & 1.54 & 2.33 & 1.33 & 1.66 & 1.99 \\
    & \cellcolor{gray!15}\textbf{\name} (Ours) & 
    \cellcolor{gray!15}\textbf{1.14} & \cellcolor{gray!15}\textbf{1.61} & \cellcolor{gray!15}\textbf{2.16} & 
    \cellcolor{gray!15}\textbf{0.61} & \cellcolor{gray!15}\textbf{1.08} & \cellcolor{gray!15}\textbf{1.61} & 
    \cellcolor{gray!15}\textbf{1.04} & \cellcolor{gray!15}\textbf{1.33} & \cellcolor{gray!15}\textbf{1.65} \\
    \midrule
    \multirow{2}{*}{$\operatorname{T\text{-}nuScenes}$} 
    & $\operatorname{SparseDrive}$~\cite{sun2024sparsedrive} & 2.07 & 2.71 & 3.36 & 0.91 & 1.71 & 2.57 & 1.54 & 2.31 & 2.90 \\
    & \cellcolor{gray!15}\textbf{\name} (Ours) & 
    \cellcolor{gray!15}\textbf{1.38} & \cellcolor{gray!15}\textbf{1.97} & \cellcolor{gray!15}\textbf{2.66} & 
    \cellcolor{gray!15}\textbf{0.47} & \cellcolor{gray!15}\textbf{0.99} & \cellcolor{gray!15}\textbf{1.66} & 
    \cellcolor{gray!15}\textbf{1.42} & \cellcolor{gray!15}\textbf{1.86} & \cellcolor{gray!15}\textbf{2.33} \\
    \bottomrule
    \end{tabular}%
    }
    \caption{ \textbf{Long trajectory planning} on the \textbf{nuScenes} and \textbf{Turning-nuScenes} validation set. 
    We train models for 10 epochs for 6s-horizon prediction.
    We follow the VAD \cite{jiang2023vad} metrics.}
    \label{table:nusc_long6}
\end{table}

\begin{table}[htp]
    \centering
    \resizebox{\linewidth}{!}{
    \begin{tabular}{l|c|c|cc} %
        \toprule
        \multirow{2}{*}{\textbf{Method}} & \multirow{2}{*}{\textbf{Input}} & \textbf{Open-loop} & \multicolumn{2}{c}{\textbf{Closed-loop Metrics}} \\
        & & Avg. L2 $\downarrow$ & Driving Score $\uparrow$ & Success Rate (\%) $\uparrow$ \\ %
        \midrule
        SparseDrive$^*$ \cite{sun2024sparsedrive} & C & 0.87 & 39.9 & 10.0 \\ %
        \gr \textbf{\name} (Ours) & C + R & \textbf{0.82} & \textbf{55.6} & \textbf{30.0} \\
        \bottomrule
    \end{tabular}}
    \caption{ 
        \textbf{Open-loop} and \textbf{closed-loop} evaluation results on \textbf{Bench2Drive} (V0.0.3) using the base training set. 
        We report the closed-loop simulation in the dev10 protocol.
        $^*$ indicates re-implementation and provided model checkpoint of \cite{song2025momad}.}
    \label{table:b2d_dev10}
\end{table}

\subsection{Limitations}
While our experiments demonstrate the benefits of radar fusion for end-to-end autonomous driving, several limitations remain. 
First, the radar data in both nuScenes and Bench2Drive provides only sparse point cloud representations, limiting the potential density of radar-based features \cite{ding2024radarocc,paek2022k,hagele2025occluded}. 
The sensing range is also restricted to 50m, which does not fully leverage radar's capability for long-range detection beyond 150m \cite{fent2024man, wolters2024sparc}. 
Additionally, the nuScenes radar setup lacks height information, preventing full 4D radar perception \cite{palffy2022multi}.
In the simulation environment of Bench2Drive, the radar sensor placement and extrinsic calibration are suboptimal compared to real-world setups. The simplified radar sensing principles in the CARLA simulator also do not fully capture the complex radar phenomenology of real sensors. 
To validate the full potential of radar-based perception for autonomous driving, extensive closed-loop testing with real-world radar-camera systems will be required.

\subsection{Future Work}
As next steps, we will explore more fusion mechanisms and extend the analysis also to dense-BEV based methods \cite{jiang2023vad, li2024hydra}.
While our current approach operates on pre-processed radar point clouds, future research directions include exploring raw radar tensor representations \cite{paek2022k, hagele2024radarcnn, fent2024dpft} and investigating larger perception ranges, potentially up to 150m \cite{fent2024man}. 
Additionally, the domain gap between simulated and real-world camera-radar data necessitates dedicated multi-modal planning-oriented datasets. 
We envision extending this work to cooperative perception scenarios \cite{yu2025end, teepe2024earlybird} on radar-camera-based V2X settings, further enhancing the robustness and safety of end-to-end autonomous driving systems \cite{xu2023v2v4real, zimmer2024tumtraf, song2024collaborative, yu2022dair}.

\section{Conclusion}
\label{sec:conclusion}

Multi-modal fusion, especially radar-based fusion, represents an overlooked yet promising research direction for end-to-end autonomous driving. 
Radar's unique characteristics -- weather robustness, Doppler velocity, and long-range capabilities beyond 150m -- enable significant improvements in scene understanding that are unavailable to vision-only approaches. 
These capabilities are highly synergistic with the overall planning requirements for safe autonomous driving.
Therefore, we introduce \textbf{\name}, a query-based camera-radar fusion framework that extends the sparse representation paradigm to planning-oriented autonomous driving. 
By integrating adaptive radar fusion strategies into a unified optimization pipeline, our approach addresses fundamental limitations of vision-centric methods, 
particularly in safety-critical scenarios where accurate motion understanding and long-horizon trajectory prediction are essential for collision avoidance.

{
    \small
    
    \bibliographystyle{ieeenat_fullname}
    \bibliography{main}
}

\end{document}